%% file: rma_ijcai25.tex
\documentclass{article}
\usepackage{ijcai25}

\usepackage{times}
\usepackage{soul}
\usepackage{url}
\usepackage[hidelinks]{hyperref}
\usepackage[utf8]{inputenc}
\usepackage[small]{caption}
\usepackage{graphicx}
\usepackage{amsmath}
\usepackage{amsthm}
\usepackage{booktabs}
\usepackage{algorithm}
\usepackage{algorithmic}

\usepackage[switch]{lineno}
\usepackage[textsize=tiny]{todonotes}
\usepackage[capitalize,noabbrev]{cleveref}
\input{math_commands.tex}

\input{utils.tex}
\usepackage{amsmath}
\usepackage{dsfont}
\usepackage{multirow}

\title{Adversarial Attacks on Both Face Recognition and Face Anti-spoofing Models}

\author{
Fengfan Zhou$^1$
\and
Qianyu Zhou$^{2,3}$\and
Hefei Ling$^1$\thanks{\textit{Corresponding author.}}\And
Xuequan Lu$^4$\\
\affiliations
$^1$School of Computer Science and Technology, Huazhong University of Science and Technology\\
$^2$College of Computer Science and Technology, Jilin University\\
$^3$ Key Laboratory of Symbolic Computation and Knowledge Engineering of Ministry of Education, JLU\\
$^4$Department of Computer Science and Software Engineering, The University of Western Australia 
\emails
\{ffzhou, lhefei\}@hust.edu.cn,
zhouqianyu@jlu.edu.cn,
bruce.lu@uwa.edu.au
}

\begin{document}

\maketitle
\begin{abstract}
Adversarial attacks on Face Recognition (FR) systems have demonstrated significant effectiveness against standalone FR models. However, their practicality diminishes in complete FR systems that incorporate Face Anti-Spoofing (FAS) models, as these models can detect and mitigate a substantial number of adversarial examples. To address this critical yet under-explored challenge, we introduce a novel attack setting that targets both FR and FAS models simultaneously, thereby enhancing the practicability of adversarial attacks on integrated FR systems. Specifically, we propose a new attack method, termed Reference-free Multi-level Alignment (RMA), designed to improve the capacity of black-box attacks on both FR and FAS models. The RMA framework is built upon three key components. Firstly, we propose an Adaptive Gradient Maintenance module to address the imbalances in gradient contributions between FR and FAS models. Secondly, we develop a Reference-free Intermediate Biasing module to improve the transferability of adversarial examples against FAS models. In addition, we introduce a Multi-level Feature Alignment module to reduce feature discrepancies at various levels of representation. Extensive experiments showcase the superiority of our proposed attack method to state-of-the-art adversarial attacks.
\end{abstract}

\section{Introduction}
\label{sec:intro}
\begin{figure*}[t]
	\begin{center}
		\centerline{\includegraphics[width=\textwidth]{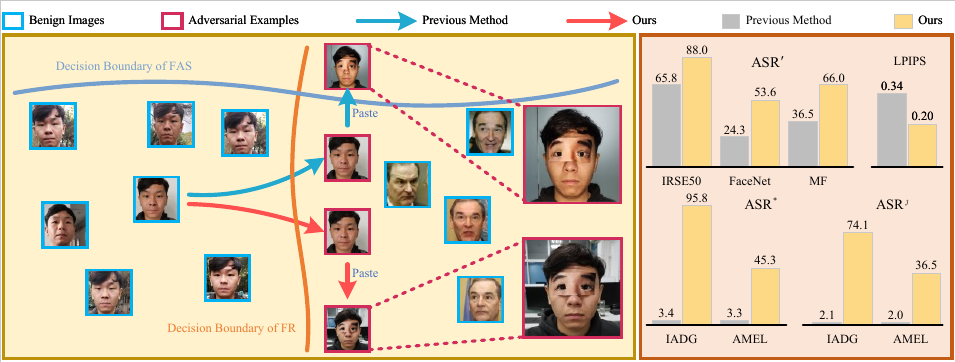}}
		\caption{
                \textbf{Left}: A comparison between previous methods and our proposed method. Adversarial examples generated by previous methods for FR systems often contain spoofing artifacts, making them easily detectable and filtered out by FAS models. In contrast, our method attacks both FR and FAS models simultaneously, improving the practicality of adversarial examples in FR systems. \textbf{Right}: Performance comparison based on the metrics of LPIPS, and Attack Success Rate (ASR) for FR (ASR$’$), FAS (ASR$^{*}$), and both models (ASR$^{\jmath}$).
		}
		\label{fig:motivation}
	\end{center}
    \vspace{-0.4cm}
\end{figure*}
Recent advancements in Face Recognition (FR) have led to remarkable performance improvements \cite{facenet,arcface,partial_fc}. However, the vulnerability of current FR systems to adversarial attacks presents a critical security concern. This underscores the urgent need to enhance the effectiveness of adversarial face examples to expose deeper vulnerabilities in FR systems. Numerous adversarial attack methods have been proposed, focusing on properties such as stealthiness \cite{semantic_adv,tip_im,low_key,amt_gan,clip2protect}, transferability \cite{dfanet,sibling_attack,bpfa,adv_restore}, and physical-world attack capability \cite{adv_makeup,at3d,cvpr23_opt_attack}. These endeavors significantly contribute to enhancing the effectiveness of adversarial attacks on FR systems.

Despite the significant progress in adversarial attacks on FR systems, the integration of Face Anti-Spoofing (FAS) \cite{zhou2022generative,zhou2024test} models in FR systems poses a substantial challenge to the practicality of these attacks in real-world scenarios. As illustrated in \cref{fig:motivation}, when adversarial examples are applied, the visual features of the source images post-pasting with adversarial examples remarkably differ from those of live images, often revealing spoof features that can be detected by FAS models. If the adversarial examples are flagged as spoof samples by FAS models, the images will be preemptively filtered out by victims without inputing into FR models, thereby hindering attempts to attack. However, existing adversarial attacks on FR often overlook the incorporation of FAS models within FR systems, leading to failures in real-world deployment scenarios. Therefore, it is imperative to develop novel adversarial attacks targeting both FR and FAS concurrently to enhance the practicality of adversarial attacks on FR systems.

To successfully attack both FR and FAS models, it is essential to utilize the gradients from both models to craft adversarial examples. However, in real-world deployment scenarios, attackers often lack direct access to the models used by their targets. Consequently, a widely adopted method is to leverage surrogate models to generate adversarial examples, subsequently transferring them to the target models~\cite{dfanet,adv_makeup,bpfa}. Nevertheless, the inherent differences between the surrogate and target models lead to disparities in their decision boundaries. As a result, when aiming to simultaneously attack FR and FAS models, it involves the following challenges.
Firstly, performing simultaneous adversarial attacks on both FR and FAS models can result in gradient imbalances, which can degrade the overall attack performance on both models.
In addition, crafting adversarial examples using the final output live score of the FAS models may lead to overfitting to the surrogate model. While employing adversarial attacks targeting the FAS models based on intermediate loss functions can enhance transferability, the resulting adversarial examples may still overfit to the specific reference live images, thereby limiting the transferability improvement.
Furthermore, generating adversarial examples for FR without considering feature alignment across multiple intermediate layers can lead to an over-reliance on specific features of the surrogate model, thereby compromising the transferability.

To address these challenges, we propose a novel attack method, termed Reference-free Multi-level Alignment (RMA), designed to simultaneously target both FR and FAS models. Our approach comprises three key modules: Adaptive Gradient Maintenance (AGM), Reference-free Intermediate Biasing (RIB), and Multi-level Feature Alignment (MFA).
We perform independent gradient calculations for the MFA and RIB modules, and subsequently the AGM module mitigates the imbalances between the two types of gradients.
Specifically, the AGM module dynamically re-weighting the losses for FR and FAS in each iteration to reduce the gradient disparities between the two tasks, thereby balancing the optimization process and enhancing the attack performance.
The RIB module biases adversarial examples into the space of the live images using an intermediate loss without overfitting to specific reference live images by approximating the neural networks leveraging the linear hypothesis from \cite{fgsm}, thereby further improving the effectiveness of black-box attacks on FAS models.
Finally, the MFA module enhances transferability across FR models by aligning the features of adversarial examples with those of target images at multiple intermediate layers of the FR models.
As depicted in \cref{fig:motivation}, after applying the adversarial examples crafted by our proposed method, the post-pasting adversarial examples can still traverse the decision boundaries of FR and FAS models, leading to successful attacks on the FR systems.
The contributions of our paper are three-fold:
\begin{itemize}
	\item We introduce a novel and practical setting of attacking FR and FAS models simultaneously to boost the practicability of adversarial attacks on FR systems. We propose an innovative adversarial attack framework termed Reference-free Multi-level Alignment (RMA) to improve the attack capacity on both models. To our best knowledge, this is the first study that adversarially attacks both FR and FAS simultaneously utilizing the gradients of both models.
        \item We design the Adaptive Gradient Maintenance module to mitigate the imbalances between the gradients of FR and FAS models, the Reference-free Intermediate Biasing module to improve the transferability of adversarial attacks on FAS, and the Multi-level Feature Alignment module to improve the black-box attack capacity of adversarial attacks on FR.
	\item Extensive experiments demonstrate the superiority of RMA to state-of-the-art adversarial attacks, as well as its nice compatibility with various models. 
\end{itemize}

\section{Related Work}
\noindent \textbf{Adversarial Attacks.}
The primary objective of adversarial attacks is to introduce imperceptible perturbations into benign images, deceiving machine learning systems and causing them to produce erroneous outputs \cite{ax_init,fgsm,rocamorarevisiting,xu2023backpropagation,shayegani2024jailbreak}. The existence of adversarial examples poses a significant threat to the security and reliability of modern machine learning systems. Consequently, substantial research efforts have been devoted to studying adversarial attacks to enhance the robustness of these systems \cite{mim,linbp,ssa6,ila_da,DBLP:conf/iccv/ChenYCCL23}.
In most real-world scenarios, attackers lack direct access to the deployed target models. To this end, numerous adversarial attacks have been introduced, aiming to boost the transferability of the adversarial examples crafted by surrogate models \cite{dim,vt,ssa6,sia,bsr}.
Despite significant progress in this field, previous adversarial attack methods often overlook improving the transferability on FAS models using intermediate loss functions without relying on specific reference images. To overcome this limitation, we leverage the linear hypothesis to approximate segments of FAS models and bias adversarial examples toward the live distribution using intermediate loss avoiding overfitting to specific reference images, thereby improving the black-box attack effectiveness on FAS models.
Additionally, the challenge of reducing feature discrepancies across various levels of representation in FR models remains largely underexplored. To address this, we propose a novel attack method that aligns the features of adversarial examples with those of target images across multiple intermediate layers, thereby enhancing transferability of the crafted adversarial examples on FR models.

\noindent \textbf{Adversarial Attacks on Face Recognition Systems.}
Adversarial attacks on FR systems can be categorized into two groups based on the constraints imposed on adversarial perturbations: restricted attacks \cite{mim,DBLP:conf/iccv/ChenYCCL23} and unrestricted attacks \cite{DBLP:conf/iccv/WeiHS023,DBLP:conf/cvpr/WeiYH23}. Restricted attacks involve generating adversarial examples within a predefined boundary, such as the $L_p$ norm constraint. The primary objective of restricted attacks is to improve the transferability of adversarial face examples across different FR models \cite{dfanet,bpfa,sibling_attack}.
In contrast, unrestricted adversarial attacks generate adversarial examples without adhering to a specific perturbation constraint. These attacks typically focus on physical-world scenarios \cite{genap,at3d,cvpr23_opt_attack}, attribute manipulation \cite{semantic_adv,adv_attribute}, or adversarial example generation through makeup transfer \cite{adv_makeup,amt_gan,clip2protect}. Both restricted and unrestricted adversarial attacks have significantly advanced the effectiveness of adversarial attacks on FR systems.
Nevertheless, in practical applications, adversarial examples generated by the methods are frequently classified as spoof images by the FAS models. This limitation significantly impacts the practicality of adversarial attacks on FR systems. In this paper, we introduce a novel attack method that simultaneously targets both FR and FAS models, thereby bolstering the effectiveness of adversarial attacks on the integrated FR systems.

\section{Methodology}
\subsection{Problem Formulation and Framework Overview}\label{sec:problem_formulation}
\noindent \textbf{Problem Formulation.}
Let $\gF^{vct}(\bfx)$ denote the FR model deployed by a victim to extract the embedding from a face image $\bfx$, and let $\gG^{vct}(\bfx)$ represent the FAS model deployed by a victim that outputs a score to determine the authenticity of the image. We denote $\xs$ and $\xt$ as the source and target images, respectively. In most cases, the source images are spoof images, and the target images are live images in real-world attack scenarios. Hence, we initialize $\xs$ with the spoof image captured in the physical world, and $\xt$ with the live image, respectively.
The objective of adversarial attacks on FR in our research is to generate an adversarial example $\xadv$ that induces the victim FR model $\gF^{vct}$ to misclassify it as the target sample $\bfx^t$, while also preserving a high level of visual similarity between $\xadv$ and $\bfx^s$.
Specifically, the objective can be stated as follows:
\begin{equation}
	\begin{gathered}
		\xadv=\mathop{\arg\min}\limits_{\xadv}\left(\mathcal{D}\left(\mathcal{F}^{vct}\left(\xadv\right), \mathcal{F}^{vct}\left(\bfx^t\right)\right)\right) \\
        \text{s.t.} \Vert \xadv - \bfx^s\Vert_p \leq \epsilon
		\label{eq:opt_obj_of_fr}
	\end{gathered}
\end{equation}
where $\mathcal{D}$ refers to a predefined distance metric, while $\epsilon$ specifies the maximum magnitude of permissible perturbation.
In contrast, the objective of the adversarial attacks on FAS in this study is to deceive the victim FAS model $\gG^{vct}$ into identifying the adversarial example $\xadv$ as a living image, while ensuring that $\xadv$ maintains a visually similar appearance from $\bfx^s$. Concisely, the objective can be formulated as:
\begin{equation}
	\begin{gathered}
		\xadv=\mathop{\arg\max}\limits_{\xadv}\left(\mathcal{G}^{vct}\left(\xadv\right)\right) \quad \text{s.t.} \Vert \xadv - \bfx^s\Vert_p \leq \epsilon
		\label{eq:opt_obj_of_fas}
	\end{gathered}
\end{equation}

\begin{figure*}[t]
	\begin{center}
		\centerline{\includegraphics[width=180mm]{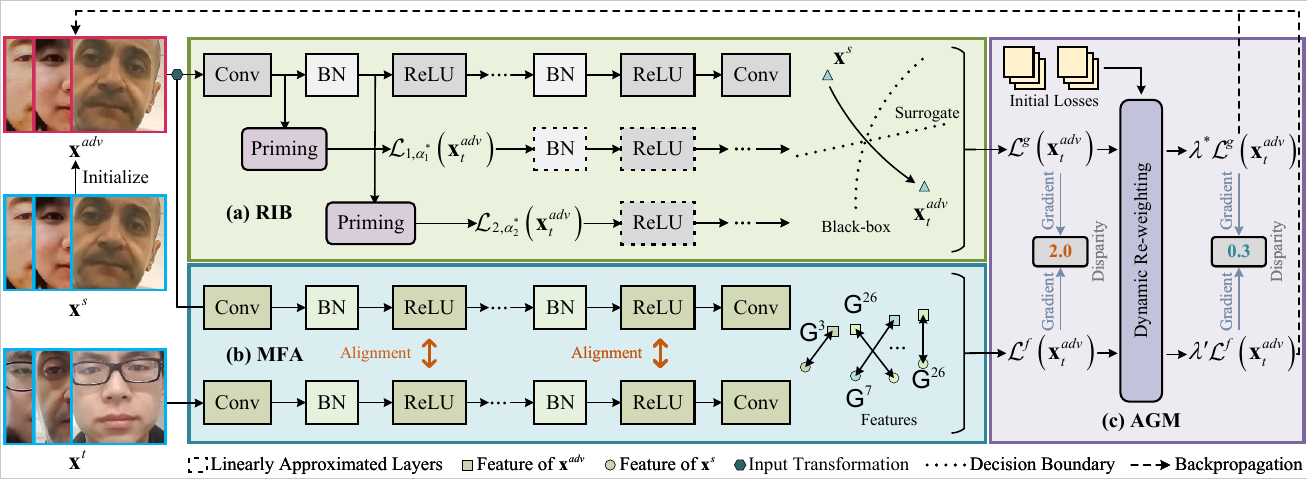}}
		\caption{
            Overview of our Reference-free Multi-level Alignment (RMA) framework.
            (a) The Reference-free Intermediate Biasing (RIB) module biases adversarial examples toward the live image space using the intermediate loss function without overfitting to specific reference live images by a surrogate model accessible to the attacker, enhancing the attack effectiveness on black-box FAS models where direct access is restricted.
            (b) The Multi-level Feature Alignment (MFA) module aligns the features of adversarial examples with those of target images across multiple intermediate layers, thereby improving their transferability when attacking FR models.
            (c) The Adaptive Gradient Maintenance (AGM) module balances the gradients between FR and FAS by adaptively scaling their respective losses, thereby mitigating the disparities between the gradients the during each iteration.
		}
		\label{fig:framework}
	\end{center}
    \vspace{-0.4cm}
\end{figure*}

\noindent \textbf{Framework Overview.}
In the realm of adversarial attacks on FR systems, the property of adversarial examples being readily identifiable by FAS models significantly hinders the practicality of such examples. To address this issue, we propose a novel attack method called Reference-free Multi-level Alignment (RMA) to attack FR and FAS models simultaneously. The overview of our proposed attack method is illustrated in \cref{fig:framework}, showcasing three key modules: Adaptive Gradient Maintenance (AGM), Reference-free Intermediate Biasing (RIB), and Multi-level Feature Alignment (MFA). 
In the following sections, we will provide a detailed introduction to each of these three modules.
\subsection{Adaptive Gradient Maintenance} 
Let $\gL^f$ and $\gL^g$ be the loss functions of the FR and FAS models, respectively. 
To craft the adversarial examples targeting both the FR and FAS tasks, a vanilla method is to compute the gradient using the following formula:
\begin{equation}
\begin{split}
\bfg'_t = \nabla_{\xadv} \gL^f \left(\xadv_t\right) + \nabla_{\xadv} \gL^g \left(\xadv_t\right)
\end{split}
\label{eq:vanilla_gradient_sum}
\end{equation}
where $\bfg'_t$ and $\xadv_t$ represent the gradient and adversarial examples at the $t$-th iteration, respectively.
However, due to the imbalances between the magnitudes of $\gL^f$ and $\gL^g$, the gradients computed for the FR and FAS models become unbalanced, which negatively impacts the performance of the adversarial examples.
To address this, we propose a novel module termed as Adaptive Gradient Maintenance (AGM), which adaptively mitigates the disparities of the gradients between FR and FAS during each iteration. 

To mitigate the gradient discrepancies between FR and FAS, we dynamically adjust the degree of loss reduction for the two tasks. Specifically, the degree of loss reduction for the FR and FAS tasks can be expressed as follows:
\begin{equation}
\begin{split}
d' = \gL^f \left(\xadv_t\right) - \gL^f \left(\xadv_1\right) \\
d^* = \gL^g \left(\xadv_t\right) - \gL^g \left(\xadv_1\right)
\end{split}
\end{equation}
After obtaining the degree of loss reduction for both tasks, we re-weight the losses of the FR and FAS models to reduce their disparities. The resulting optimization objective is as follows:
\begin{equation}
\begin{split}
\xadv &= \arg \max \gL^f \left(\xadv\right)+ \frac{d'}{d^*}\gL^g\left(\xadv\right) \\
&= \arg \max \lambda' \gL^f \left(\xadv\right)+ \lambda^*\gL^g\left(\xadv\right)
\end{split}
\label{eq:agm_opt_obj}
\end{equation}
where $\lambda'$ and $\lambda^*$ are the normalized loss weights:
\begin{equation}
\begin{split}
\lambda' = \frac{d^*}{d^* + d' + \epsilon'}, \quad \lambda^* = \frac{d'}{d^* + d' + \epsilon'}
\end{split}
\end{equation}
Here, $\epsilon'$ is a small constant to ensure numerical stability.

Using the optimization objective in the equation above, we compute the balanced gradient $\bfg$ using the following formula by dynamically re-weighting the loss magnitude:
\begin{equation}
\begin{split}
\bfg_t = \nabla_{\xadv} \lambda' \gL^f \left(\xadv_t\right) + \nabla_{\xadv} \lambda^* \gL^g \left(\xadv_t\right)
\end{split}
\label{eq:balanced_gradient}
\end{equation}
After obtaining $\bfg_t$, we use $\bfg_t$ to update the adversarial examples according to the following formula:
\begin{equation}
\begin{split}
\xadv_{t+1} = \prod \limits_{\xs, \epsilon}\left(\xadv_{t}- {\rm sign}\left(\bfg_{t}\right)\right)
\end{split}
\end{equation}
\subsection{Reference-free Intermediate Biasing} \label{sec:rib}
Unlike FR, which is a metric learning task, the FAS task involves the model directly producing a score to determine the liveness of the input face image. Once the liveness score is obtained, it is compared against a predefined threshold. If the score exceeds the threshold, the input face image is classified as a live image, otherwise a spoof image.

Let \(\gG\) denote the FAS surrogate model. The optimization objective of the attack on FAS using $\gG$ can be expressed as:
\begin{equation}
	\max_{\xadv} \gG\left(\xadv\right)  \quad {\rm s.t.} \quad \Vert \xadv - \bfx^s\Vert_p \leq \epsilon\label{eq:opt_obj_fas}
\end{equation}
where $\epsilon$ specifies the maximum perturbation magnitude.

A vanilla method to achieve \cref{eq:opt_obj_fas} is to utilize the following loss function:
\begin{equation}
	\gL^s\left(\xadv\right)=-\gG\left(\xadv\right) \label{eq:l_s}
\end{equation}
where $\gL^s$ denotes the loss function for generating adversarial examples against FAS models using the method.
\cref{eq:l_s} is effective in crafting adversarial examples for FAS. However, as demonstrated by previous methods~\cite{adv_attribute}, using a loss function based on intermediate layers to craft adversarial examples can further improve transferability. Although these adversarial attacks which leverage intermediate layer loss functions demonstrate promising effectiveness, no such attack has been specifically designed to target FAS models to the best of our knowledge.

For clarity, we focus on models with a single branch in their computational graphs. For models with multiple branches, our RMA remains largely unchanged, with minor modifications to accommodate the handling of multiple branches.
Let \(g^i\) denote the \(i\)-th layer of the model \(\gG\) and let $l$ represent the total number of layers in \(\gG\).
We define the segment of \(\gG\) from layer \(g^i\) to layer \(g^j\) as follows \cite{dpa}:
\begin{equation}
	\gG^{i,j} = g^i \circ g^{i+1} \circ \cdots \circ g^{j-1} \circ g^j, \label{eq:network_segment}
\end{equation}
where \(\circ\) denotes the composition of functions operation.
Let \(\sfG^k(\bfx)\) represent \(\gG^{1,k}(\bfx)\). To craft adversarial examples for FAS models using an intermediate layer loss function, a straightforward method is to use the following loss function:
\begin{equation}
	\gL_k' \left(\xadv\right) = \left \Vert \sfG^k \left(\xadv\right) - \sfG^k \left(\bfx^*\right)\right \Vert_2 \label{eq:l_d}
\end{equation}
where $\bfx^*$ denotes a pre-selected reference live image. However, the adversarial examples crafted by \cref{eq:l_d} may overfit to \(\bfx^*\). If adversarial examples for FAS can be crafted using an intermediate layer loss function without relying on specific reference live images, better transferability can be achieved (See \cref{fig:ablation_studies_rib}).

However, leveraging \(\sfG^k(\xadv)\) solely to maximize the live score produced by \(\gG\) in the final outputs poses a significant challenge due to the absence of the $\gG^{k+1,l}$. To address this, we introduce the linear hypothesis proposed by \cite{fgsm}. Let us revisit an early hypothesis posited by \cite{fgsm}, which suggests that the linear nature of modern deep neural networks, resembling linear models trained on the same dataset, is the underlying cause of adversarial examples and their surprising transferability \cite{linbp,adv_pruning}. Let \( m_k \) denote the number of units in \(\sfG^k(\xadv)\). Based on this hypothesis, the FAS model \(\gG^{k+1,l}\) can be linearly approximated as:
\begin{equation}
	\gG^{k+1,l} \left(\sfG^k \left(\xadv\right)\right)=\bfw^{k+1,l} \Psi\left( \sfG^k \left(\xadv\right)\right)^\top + \bfb^{k+1,l}\label{eq:g_linear_form}
\end{equation}
where $\Psi$ denotes the flatten operation, \(\bfw^{k+1,l} \in \bbR^{1 \times m_k}\) is the weight vector, and \(\bfb^{k+1,l} \in \bbR^{1 \times 1}\) is the bias term.
Let $\bfh^k$ be defined as $\bfh^k=\Psi \left(\sfG^k \left(\xadv\right) \right)^\top \in \bbR^{m_k \times 1}$. Based on \cref{eq:g_linear_form}, the objective for the FAS model can be expressed as the following formula:
\begin{equation}
	\max_{\xadv} \gG\left(\xadv\right)=\max_{\xadv} \bfw^{k+1,l} \bfh^k\label{eq:obj_linear_form}
\end{equation}
Let $t$ be the number of the optimization iterations. Since $\bfw^{k+1,l}$ is intractable, we cannot use $-\bfw^{k+1,l} \bfh^k$ as the loss function to optimize \cref{eq:obj_linear_form}. Instead, we design an optimization process that ensures the following condition:
\begin{equation}
\begin{split}
    \bfw^{k+1,l} \bfh^k_t - \bfw^{k+1,l} \bfh^k_{t-1}=\bfw^{k+1,l} \left(\bfh^k_t - \bfh^k_{t-1}\right) >0
\end{split}
\label{eq:wb_guarantee}
\end{equation}
where $\bfh^k_t$ denotes the value of $\bfh^k$ at the $t$-th optimization iteration.
In our research, the parameters of the FAS model are fixed, meaning that $\bfw^{k+1,l}$ is a vector with constant values.
Therefore, we opt to satisfy \cref{eq:wb_guarantee} by minimizing or maximizing $\bfh^k$ using the following loss function:
\begin{equation}
\begin{split}
    \widehat{\gL}_{k,\alpha_k}\left(\xadv\right) =\frac{\alpha_k}{m_k} \sum_{j=0}^{m_k} \bfh^{k,j} \quad {\rm s.t} \quad \alpha_k \in \left\{-1, 1\right\}
\end{split}
\label{eq:widehat_l}
\end{equation}
where $\bfh^{k,j}$ denotes the $j$-th element in $\bfh^k$.
The value of $\alpha_k$ varies across different layers depending on layer index $k$. To determine $\alpha_k$, we introduce a stage termed as Prime, which records the value of $\gL^s$ obtained by crafting adversarial examples using \cref{eq:widehat_l} with both candidate $\alpha_k$ values (\textit{i.e.} -1 or 1). The $\alpha_k$ value corresponding to the lower $\gL^s$ is then selected as the final $\alpha_k$. 
Specifically, let $\bfx'$ represent the adversarial examples generated during the Priming stage, initialized with the same values as $\xs$. Note that $\bfx'$ is only used to calculate the optimization direction. After the Priming stage, $\bfx'$ is discarded and is not involved in the final adversarial example generation process. $\bfx'$ is optimized over $b$ iterations using the following formula~\cite{i_fgsm}:
    \begin{equation}
    \begin{split}
    \bfx'_{\alpha_k, t} = \prod \limits_{\xs, \epsilon}\left(\bfx'_{\alpha_k, t-1}- {\rm sign}\left(\nabla_{\bfx'_{\alpha_k, t-1}}\widehat{\gL}_{k, \alpha_k}\left(\bfx'_{\alpha_k, t-1}\right)\right)\right)
    \end{split}
    \end{equation}
where $\bfx'_{\alpha_k, t}$ denotes the adversarial example crafted in the $t$-th iteration using $\alpha_k$ to compute the loss function, $\prod \limits_{\xs, \epsilon}$ represents the clipping operation that ensures the distance between the crafted adversarial example $\bfx'$ and the source image $\xs$ remains within $\epsilon$. During the optimization process, we record the $\gL^s$ loss using \cref{eq:l_s} and compute the average loss values as follows:
\begin{equation}
\bar{\gL}_{k, \alpha_k} = \frac{1}{b}\sum_{t=1}^{b} \gL^s\left(\bfx'_{\alpha_k, t}\right)
\label{eq:widebar_l}
\end{equation}
Using \cref{eq:widebar_l}, we determine the optimal $\alpha_k$ for $\widehat{\gL}_{k,\alpha_k}$ by the following formula:
\begin{equation}
\alpha_k^{*}=\arg \min \bar{\gL}_{k, \alpha_k}
\end{equation}

Note that if the layer is the final output layer of the FAS model, we can directly judge the $\alpha_k$ value of $\widehat{\gL}_{k,\alpha_k}$ and do not need to process the Prime stage to calculate it.
We use the loss in multiple layers to craft the adversarial examples:
\begin{equation}
\begin{split}
\gL^g \left(\xadv\right) = \sum_{i=1}^{\left \vert s \right \vert} \widehat{\gL}_{s_i,\alpha_{s_i}^*}\left(\xadv\right)
\end{split}
\label{eq:l_g}
\end{equation}
where $s$ is the pre-defined layer index set to calculate the loss for the FAS task.

\subsection{Multi-level Feature Alignment} \label{sec:mfa}
Let $\gF$ represent the FR surrogate model.
To deceive FR models, a vanilla method is to utilize the following loss function to generate adversarial examples~\cite{dfanet,bpfa}:
\begin{equation}
	\gL^{i}\left(\xadv\right)=\left\Vert \phi\left(\gF\left(\xadv\right)\right)-\phi\left(\gF\left(\xt\right)\right)\right\Vert^2_2\label{eq:l_i}
\end{equation}
where $\phi(\bfx)$ represents the operation that normalizes $\bfx$.
However, if we rely solely on \cref{eq:l_i} to craft adversarial face examples, the attack success rate will be limited, as only the final feature of is utilized. 

To improve the transferability on FR models, we propose the Multi-level Feature Alignment (MFA) module. MFA enhances the transferability of adversarial attacks by aligning the features of adversarial examples with those of target images across multiple intermediate layers.
To the best of our knowledge, MFA is the first algorithm to employ a multi-level intermediate loss function for enhancing the transferability of the adversarial examples.

Let \( e \) denote the index set of pre-selected layers used for calculating the loss to align the intermediate features, and let \(\sfF^k(\bfx)\) represent the feature of the $k$-th layer.
For each layer specified by \( e \), we use the following formula to compute the loss, which aims to align the features across different levels of the intermediate layers:
\begin{equation}
    \gL^f\left(\xadv\right)=\sum_i^{\left \vert e \right \vert} \left\Vert \phi\left(\Psi\left(\sfF^{e_i}\left(\xadv\right)\right)\right) -\phi\left(\Psi\left(\sfF^{e_i}\left(\xt\right)\right)\right)\right\Vert^2_2 \label{eq:l_f}
\end{equation}

\begin{table*}[!htbp]
\centering
\begin{tabular}{c|ccc|cc|cccccc}
\hline
\multicolumn{1}{l|}{} & \multicolumn{3}{c|}{ASR$^\prime$}                                                       & \multicolumn{2}{c|}{ASR$^*$}                       & \multicolumn{6}{c}{ASR$^{\jmath}$}                                                                                                                                                                           \\ \hline
Attacks               & \multicolumn{1}{c|}{IR152}         & \multicolumn{1}{c|}{IRSE50}        & FaceNet       & \multicolumn{1}{c|}{IADG}          & AMEL          & \multicolumn{1}{c|}{IR152$^\prime$} & \multicolumn{1}{c|}{IRSE50$^\prime$} & \multicolumn{1}{c|}{FaceNet$^\prime$} & \multicolumn{1}{c|}{IR152$^*$}     & \multicolumn{1}{c|}{IRSE50$^*$}    & FaceNet$^*$   \\ \hline
FIM                   & \multicolumn{1}{c|}{23.5}          & \multicolumn{1}{c|}{20.9}          & 25.0          & \multicolumn{1}{c|}{5.3}           & 3.0           & \multicolumn{1}{c|}{1.1}            & \multicolumn{1}{c|}{1.1}             & \multicolumn{1}{c|}{1.5}              & \multicolumn{1}{c|}{1.0}           & \multicolumn{1}{c|}{0.9}           & 0.9           \\
DI                    & \multicolumn{1}{c|}{34.5}          & \multicolumn{1}{c|}{32.6}          & 41.5          & \multicolumn{1}{c|}{3.3}           & 2.3           & \multicolumn{1}{c|}{1.2}            & \multicolumn{1}{c|}{1.0}             & \multicolumn{1}{c|}{1.4}              & \multicolumn{1}{c|}{1.0}           & \multicolumn{1}{c|}{0.7}           & 1.2           \\
DFANet                & \multicolumn{1}{c|}{30.9}          & \multicolumn{1}{c|}{26.7}          & 32.2          & \multicolumn{1}{c|}{5.2}           & 2.4           & \multicolumn{1}{c|}{1.5}            & \multicolumn{1}{c|}{1.5}             & \multicolumn{1}{c|}{1.5}              & \multicolumn{1}{c|}{1.0}           & \multicolumn{1}{c|}{0.9}           & 1.0           \\
VMI                   & \multicolumn{1}{c|}{36.3}          & \multicolumn{1}{c|}{29.2}          & 34.8          & \multicolumn{1}{c|}{2.3}           & 1.9           & \multicolumn{1}{c|}{1.0}            & \multicolumn{1}{c|}{0.7}             & \multicolumn{1}{c|}{0.8}              & \multicolumn{1}{c|}{0.8}           & \multicolumn{1}{c|}{0.7}           & 0.7           \\
SSA                   & \multicolumn{1}{c|}{31.8}          & \multicolumn{1}{c|}{27.1}          & 35.4          & \multicolumn{1}{c|}{4.0}           & 2.5           & \multicolumn{1}{c|}{1.3}            & \multicolumn{1}{c|}{1.3}             & \multicolumn{1}{c|}{1.2}              & \multicolumn{1}{c|}{0.9}           & \multicolumn{1}{c|}{1.2}           & 1.0           \\
SIA                   & \multicolumn{1}{c|}{38.9}          & \multicolumn{1}{c|}{33.3}          & 43.8          & \multicolumn{1}{c|}{2.3}           & 1.5           & \multicolumn{1}{c|}{1.1}            & \multicolumn{1}{c|}{0.9}             & \multicolumn{1}{c|}{1.0}              & \multicolumn{1}{c|}{0.6}           & \multicolumn{1}{c|}{0.7}           & 0.7           \\
BPFA                  & \multicolumn{1}{c|}{34.0}          & \multicolumn{1}{c|}{31.4}          & 37.3          & \multicolumn{1}{c|}{3.8}           & 2.3           & \multicolumn{1}{c|}{1.5}            & \multicolumn{1}{c|}{1.3}             & \multicolumn{1}{c|}{1.4}              & \multicolumn{1}{c|}{0.9}           & \multicolumn{1}{c|}{1.1}           & 1.0           \\
BSR                   & \multicolumn{1}{c|}{25.8}          & \multicolumn{1}{c|}{22.8}          & 36.4          & \multicolumn{1}{c|}{2.1}           & 1.5           & \multicolumn{1}{c|}{0.5}            & \multicolumn{1}{c|}{0.5}             & \multicolumn{1}{c|}{0.6}              & \multicolumn{1}{c|}{0.6}           & \multicolumn{1}{c|}{0.6}           & 0.5           \\
Ours                  & \multicolumn{1}{c|}{\textbf{61.3}} & \multicolumn{1}{c|}{\textbf{55.7}} & \textbf{57.5} & \multicolumn{1}{c|}{\textbf{94.4}} & \textbf{41.0} & \multicolumn{1}{c|}{\textbf{58.7}}  & \multicolumn{1}{c|}{\textbf{53.2}}   & \multicolumn{1}{c|}{\textbf{55.6}}    & \multicolumn{1}{c|}{\textbf{29.1}} & \multicolumn{1}{c|}{\textbf{27.0}} & \textbf{26.8} \\ \hline
\end{tabular}
\caption{Comparisons of ASR (\%) results for adversarial attacks.}
\label{tab:compare_study_oulu_npu_mf}
\end{table*}

\section{Experiment}
In our experiments, we demonstrate the superiority and key properties of the proposed method. \cref{sec:experimental_setting} details the experimental settings, while \cref{sec:comparison_study} presents the comparative results. Additionally, \cref{sec:ablation_study} provides an analysis of the ablation studies.
\subsection{Experimental Settings}\label{sec:experimental_setting}
We use the Oulu-NPU \cite{oulu_npu} and CASIA-MFSD \cite{casia_mfsd} for evaluation. We randomly sample 1,000 negative image pairs from both datasets. To align with practical attack scenarios, we select spoof images captured by cameras in the physical-world as the source images and live images as the target images. We employ the Attack Success Rate (ASR) as the primary metric to evaluate the effectiveness of adversarial examples. Following previous works such as \cite{adv_makeup,amt_gan,bpfa,adv_pruning}, we selected IR152 \cite{resnet}, IRSE50 \cite{irse50}, FaceNet \cite{facenet}, and MobileFace (abbreviated as MF) \cite{arcface} as the models to assess the attack performance on FR.
we selected restricted attacks on FR systems DFANet~\cite{dfanet} and BPFA~\cite{bpfa}, as well as state-of-the-art transfer attacks DI~\cite{dim}, SSA~\cite{ssa6}, SIA~\cite{sia}, and BSR~\cite{bsr}, as our baseline for comparison.

\begin{figure}[t]
	\begin{center}
		\centerline{\includegraphics[width=\columnwidth]{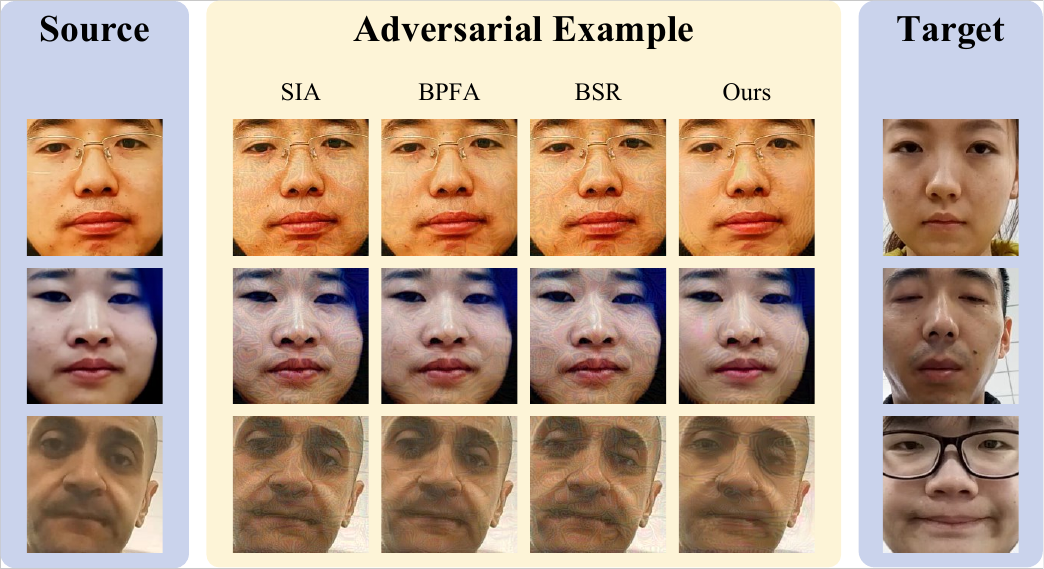}}
		\caption{
                The illustration of the crafted adversarial examples.}
		\label{fig:ax_show}
	\end{center}
    \vspace{-0.6cm}
\end{figure}

\subsection{Comparison Studies}\label{sec:comparison_study}
\noindent \textbf{RMA achieves the best black-box attack results across various ASR metrics.}
Let \(\sfN\) denote the name of a target FR model. In the following, we use \(\sfN^\prime\) and \(\sfN^*\) to represent the ASR\(^{\jmath}\) results on the IADG and AMEL models, respectively. We employ MF as the surrogate model, and craft adversarial examples on the OULU-NPU. The results are presented in \cref{tab:compare_study_oulu_npu_mf}. Some of the crafted adversarial examples are demonstrated in \cref{fig:ax_show}. These results demonstrate that our proposed method significantly outperforms previous attack methods across the ASR\('\), ASR\(^{*}\), and ASR\(^{\jmath}\) metrics, underscoring the effectiveness of our proposed attack method.

\noindent \textbf{RMA delivers superior black-box performance on adversarial robust models.}
In practical scenarios, victims may employ adversarial robust models to defend against adversarial attacks. Therefore, assessing the effectiveness of adversarial attacks on robust models is crucial. In this study, we generate adversarial examples using MF as the surrogate model and evaluate the performance of various attacks on adversarial robust models. The results, presented in \cref{tab:adv_robust_models}, demonstrate that our proposed method significantly outperforms baseline adversarial attacks. These findings underscore the effectiveness of our approach against adversarial robust models.

\begin{table}[t]
\centering
\small
\begin{tabular}{c|ccc|c}
\hline
\multicolumn{1}{l|}{} & \multicolumn{3}{c|}{ASR$^\prime$}                                                          & ASR$^{\jmath}$ \\ \hline
Attacks               & \multicolumn{1}{c|}{IR152$^{adv}$} & \multicolumn{1}{c|}{IRSE50$^{adv}$} & FaceNet$^{adv}$ & Average        \\ \hline
FIM                   & \multicolumn{1}{c|}{27.4}          & \multicolumn{1}{c|}{35.9}           & 31.8            & 1.4            \\
DI                    & \multicolumn{1}{c|}{43.9}          & \multicolumn{1}{c|}{53.5}           & 49.5            & 1.8            \\
DFANet                & \multicolumn{1}{c|}{35.5}          & \multicolumn{1}{c|}{43.9}           & 37.7            & 1.4            \\
VMI                   & \multicolumn{1}{c|}{43.0}          & \multicolumn{1}{c|}{52.9}           & 41.6            & 1.0            \\
SSA                   & \multicolumn{1}{c|}{40.3}          & \multicolumn{1}{c|}{49.3}           & 44.5            & 1.6            \\
SIA                   & \multicolumn{1}{c|}{45.9}          & \multicolumn{1}{c|}{55.3}           & 52.5            & 1.1            \\
BPFA                  & \multicolumn{1}{c|}{41.2}          & \multicolumn{1}{c|}{48.6}           & 45.2            & 1.7            \\
BSR                   & \multicolumn{1}{c|}{31.1}          & \multicolumn{1}{c|}{41.0}           & 44.7            & 0.8            \\
Ours                  & \multicolumn{1}{c|}{\textbf{70.9}} & \multicolumn{1}{c|}{\textbf{74.0}}  & \textbf{63.3}   & \textbf{49.3}  \\ \hline
\end{tabular}
\caption{ASR (\%) results on adversarial robust models.}
\label{tab:adv_robust_models}
\vspace{-0.6cm}
\end{table}
\subsection{Ablation Studies}\label{sec:ablation_study}
\noindent \textbf{The individual contributions of each module in our proposed method.}
In this section, we conduct ablation studies to evaluate the individual contributions of each module in our proposed method. Using FIM as the baseline, we incrementally integrate each module and analyze their impact. The ablation studies are performed using MF as surrogate models on the OULU-NPU dataset. We evaluate the ASR\(^{\jmath}\) results with IR152, IRSE50, and FaceNet as the target FR models and AMEL as the target FAS model. The results are presented in \cref{tab:ablation_study}.
The baseline ASR\(^{\jmath}\) values are 1.0\%, 0.9\%, and 0.9\% for IR152, IRSE50, and FaceNet, respectively. By incorporating the MFA module, the performance improves to 1.2\%, 1.5\%, and 1.2\%, respectively. The addition of the RIB module further enhances the results, achieving ASR\(^{\jmath}\) values of 14.7\%, 14.0\%, and 13.7\% for IR152, IRSE50, and FaceNet, respectively. Finally, integrating the AGM module significantly increases the ASR\(^{\jmath}\) to 29.1\%, 27.0\%, and 26.8\% for IR152, IRSE50, and FaceNet, respectively. These substantial improvements demonstrate the effectiveness of each module in our proposed method, collectively contributing to a significant enhancement in overall performance.
\begin{table}[!htbp]
\centering
\begin{tabular}{ccc|c|c|c}
\hline
MFA        & RIB        & AGM        & IR152 & IRSE50 & FaceNet \\ \hline
-          & -          & -          & 1.0            & 0.9             & 0.9              \\
\checkmark & -          & -          & 1.2& 1.5& 1.2\\
\checkmark & \checkmark & -          & 14.7& 14.0& 13.7\\
\checkmark & \checkmark & \checkmark & \textbf{29.1}& \textbf{27.0}& \textbf{26.8}\\ \hline
\end{tabular}
\caption{Comparisons of ASR$^{\jmath}$ (\%) results with AMEL as the target model on the OULU-NPU dataset.}
\label{tab:ablation_study}
\vspace{-0.2cm}
\end{table}

\begin{figure}[!htbp]
	\begin{center}
		\centerline{\includegraphics[width=\textwidth/2]{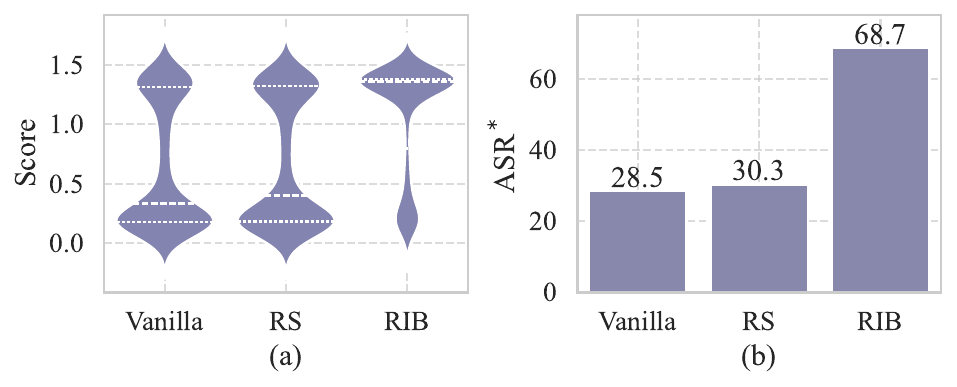}}
		\caption{
                Ablation studies on the RIB module: (a) Violin plot illustrating the black-box live scores. (b) ASR$^*$ (\%) results.
                }
		\label{fig:ablation_studies_rib}
	\end{center}
    \vspace{-0.6cm}
\end{figure}

\noindent \textbf{The Effectiveness of Reference-free Intermediate Biasing.}
To evaluate the effectiveness of our proposed RIB module, we conduct experiments using IADG as the surrogate model and AMEL as the target model on the OULU-NPU dataset. First, we craft adversarial examples using $\gL^s$, referring to this attack method as Vanilla. Second, we craft adversarial examples using $\gL'_k$, denoting this method as Reference-specific (abbreviated as RS). Finally, we craft adversarial examples using $\gL^g$, referring to this method as RIB. We evaluate the live score of the AMEL model and the ASR$^*$ performance on the AMEL model for the three attack methods. The results for these attack methods are illustrated in \cref{fig:ablation_studies_rib}.

The left plot in \cref{fig:ablation_studies_rib} shows that the proportion of high live scores achieved by RIB is greater than that of Vanilla and RS, indicating that adversarial examples crafted by our proposed RIB module are more likely to succeed in attacks. The right plot in \cref{fig:ablation_studies_rib} demonstrates that the ASR$^*$ results of our proposed RIB method surpass those of Vanilla and RS, further validating the effectiveness of the RIB module.

\noindent \textbf{The Effectiveness of Multi-level Feature Alignment.}
To evaluate the effectiveness of our proposed Multi-level Feature Alignment (MFA) module, we conduct a series of experiments using MF as the surrogate model on the OULU-NPU dataset. First, we craft adversarial examples using $\gL^i$ and refer to this attack method as Vanilla. Second, we generate adversarial examples using the following loss function, which operates on a single intermediate layer:
\begin{equation}
    \widetilde{\gL}\left(\xadv\right)=\left\Vert \phi\left(\Psi\left(\sfF^r\left(\xadv\right)\right)\right) -\phi\left(\Psi\left(\sfF^r\left(\xt\right)\right)\right)\right\Vert^2_2 \label{eq:l_f}
\end{equation}
where $r$ denote the index of the pre-defined layer for calculating the loss. We denote this attack method as Single-level. Finally, we craft the adversarial examples using $\gL^f$, referring to this attack method as MFA. The results of these three attack methods are presented in \cref{tab:ablation_study_mfa}.

\begin{table}[!htbp]
\centering
\begin{tabular}{c|cccc}
\hline
             & IR152         & IRSE50        & FaceNet       & ASR$^{adv}$   \\ \hline
Vanilla      & 23.5          & 20.9          & 25.0          & 31.7          \\
Single-level & 34.2          & 33.9          & 27.9          & 35.7          \\
MFA (Ours)   & \textbf{62.3} & \textbf{58.4} & \textbf{45.8} & \textbf{63.3} \\ \hline
\end{tabular}

\caption{Ablation study results (ASR$^\prime$, \%) for MFA. ASR$^{adv}$ is the average black-box ASR$^\prime$ on IR152$^{adv}$, IRSE50$^{adv}$, and FaceNet$^{adv}$.}
\label{tab:ablation_study_mfa}
\end{table}
\cref{tab:ablation_study_mfa} demonstrates that although the performance of Single-level surpasses that of Vanilla, it remains inferior to MFA, highlighting the effectiveness of the multi-level intermediate loss and the proposed MFA module.

\section{Conclusion}
In this paper, we introduce a novel and practical setting that aims to simultaneously attack both the FR and FAS models, thereby enhancing the practicability of adversarial attacks on FR systems. To achieve this goal, we propose an innovative framework RMA, consisting of three modules: Adaptive Gradient Maintenance (AGM), Reference-free Intermediate Biasing (RIB), and Multi-level Feature Alignment (MFA) modules.
Firstly, to alleviate the unbalance between the gradients of the FR and FAS models, we introduce the Adaptive Gradient Maintenance module, which balances the gradients on FR and FAS by adaptively re-weighting the loss on FR and FAS to decrease their disparity in each iteration.
Furthermore, to enhance the transferability of FAS models, we design the Reference-free Intermediate Biasing module to bias the adversarial examples into the space of the live image using intermediate loss without overfiting to specific reference live images.
In addition, the Multi-level Feature Alignment module is proposed to boost the capacity of black-box attacks on FR models.
Extensive experiments demonstrate the effectiveness of our proposed attack method.

\section{Acknowledgment}
This work was supported in part by the Natural Science Foundation of China under Grant 62372203 and 62302186, in part by the Major Scientific and Technological Project of Shenzhen (202316021), in part by the National key research and development program of China (2022YFB2601802), in part by the Major Scientific and Technological Project of Hubei Province (2022BAA046, 2022BAA042).

\bibliographystyle{named}
\bibliography{ijcai25}
\end{document}

%% file: math_commands.tex
\usepackage{amsmath,amsfonts,bm}

\def\eqref#1{equation~\ref{#1}}

\def\1{\bm{1}}

\def\bfb{{\mathbf{b}}}

\def\bfg{{\mathbf{g}}}
\def\bfh{{\mathbf{h}}}

\def\bfw{{\mathbf{w}}}
\def\bfx{{\mathbf{x}}}

\DeclareMathAlphabet{\mathsfit}{\encodingdefault}{\sfdefault}{m}{sl}
\SetMathAlphabet{\mathsfit}{bold}{\encodingdefault}{\sfdefault}{bx}{n}

\def\gF{{\mathcal{F}}}
\def\gG{{\mathcal{G}}}

\def\gL{{\mathcal{L}}}

\newcommand{\xadv}{\mathbf{x}^{adv}}
\newcommand{\xs}{\mathbf{x}^{s}}
\newcommand{\xt}{\mathbf{x}^{t}}

\def\sfF{{\mathsf{F}}}
\def\sfG{{\mathsf{G}}}

\def\sfN{{\mathsf{N}}}

\def\bbR{{\mathbb{R}}}